\documentclass[journal]{IEEEtran}
\usepackage{caption}
\usepackage{graphicx}
\usepackage{cite}
\usepackage{picinpar}
\usepackage{amsmath,bm}
\usepackage{url}
\usepackage{flushend}
\usepackage[latin1]{inputenc}
\usepackage{colortbl}
\usepackage{soul}
\usepackage{multirow}
\usepackage{pifont}
\usepackage{color}
\usepackage{alltt}
\usepackage[hidelinks]{hyperref}
\usepackage{enumerate}
\usepackage{siunitx}
\usepackage{epstopdf}
\usepackage{pbox}
\usepackage{algorithm,algorithmic}
\newtheorem{remark}{Remark}
\usepackage{subfigure}
\usepackage{amssymb}
\usepackage{epsfig}
\usepackage{makecell} 
\usepackage{booktabs}
\usepackage{stfloats}

\begin{document}

\title{	Fuzzy Explainable Continual Hierarchical Reinforcement Learning for Robotic Navigation}

\author
{ Peng Qin, Tao Zhao,

\thanks{Peng Qin,(qinpengdeyouxiang@outlook.com).}
\thanks{Tao Zhao,(zhaotaozhaogang@126.com).}

}
\maketitle

\begin{abstract}
Continual robot navigation requires reinforcement learning policies to continuously adapt to new environments while retaining previously acquired knowledge. Meanwhile, explainable reinforcement learning aims to learn meaningful behavioral representations through policy interaction. However, these objectives are inherently conflicting, as continual learning emphasizes policy stability whereas explainable behavior learning requires adaptive exploration. Existing methods rarely address both simultaneously. To resolve this issue, this paper proposes a Fuzzy Explainable Continual Hierarchical Reinforcement Learning (FECHRL) method. A two-stage behavior cloning strategy is introduced to balance rapid policy adaptation and explainable behavior evolution by progressively transitioning from strong imitation guidance to autonomous reinforcement learning. Furthermore, a fuzzy critic based on a Takagi--Sugeno fuzzy inference system is proposed to alleviate value extrapolation errors during the transition between imitation-guided learning and autonomous exploration, thereby improving continual training stability. Experimental results on sequential multi-scenario navigation tasks demonstrate that the proposed method effectively unifies continual learning capability and policy explainability within a single reinforcement learning framework, achieving stable knowledge retention and meaningful behavioral explanation.
\end{abstract}

\markboth{}%
{}

\definecolor{limegreen}{rgb}{0.2, 0.8, 0.2}
\definecolor{forestgreen}{rgb}{0.13, 0.55, 0.13}
\definecolor{greenhtml}{rgb}{0.0, 0.5, 0.0}


\section{Introduction}
\IEEEPARstart{N}avigation is one of the fundamental capabilities required for autonomous robots operating in complex environments, with broad applications in transportation, inspection, and service robotics. In recent years, reinforcement learning (RL) has demonstrated strong decision-making capability in end-to-end navigation tasks. However, robotic navigation in continual multi-environment scenarios still faces several important challenges.

First, when learning new navigation tasks, RL often requires extensive interaction and random exploration, resulting in low learning efficiency and unstable policy optimization in complex environments. Although prior knowledge and behavior cloning (BC) can alleviate this issue by accelerating policy initialization, excessive dependence on cloning guidance may constrain policy adaptability and reduce the capability of autonomous behavior learning. Second, in continual learning (CL) scenarios, adapting to new environments often causes catastrophic forgetting of previously learned navigation skills, thereby degrading long-term task performance. Furthermore, most existing RL-based navigation methods rely on black-box policy representations, making it difficult to provide explainability for navigation behaviors and internal decision-making processes.

Existing studies have attempted to address these problems from different perspectives. Continual reinforcement learning (CRL) methods alleviate forgetting through parameter regularization and memory replay mechanisms\cite{ewc,SCDMSLL}, while explainable RL methods improve policy explainability through hierarchical structures, latent behavior representations, or reward decomposition\cite{21,22}. In addition, BC and imitation learning have been widely used to accelerate policy learning and reduce negative transfer during early-stage training. However, these approaches are typically studied independently. In particular, CL and explainability-oriented policy learning exhibit conflicting requirements. CL prefers stronger prior constraints to stabilize policy transfer, whereas explainability-oriented behavior learning requires sufficient autonomous interaction to form environment-adaptive and semantically meaningful behavior patterns. Excessive cloning guidance may therefore suppress the emergence of explainable behavior representations.

To address this issue, this study proposes an Explainable Fuzzy-Enhanced Continual Hierarchical Reinforcement Learning method, which jointly considers CL capability and policy explainability within a unified navigation framework. To balance knowledge guidance and autonomous behavior formation, a two-stage behavior cloning mechanism is introduced. During the early training stage, strong cloning supervision is employed to accelerate policy acquisition and alleviate negative transfer, while in the later stage the cloning constraint is weakened to release exploration capability and enable the formation of explainable behavior patterns through environmental interaction. Furthermore, to stabilize the transition from cloning-guided learning to autonomous interaction learning, a  fuzzy critic is proposed to reduce extrapolation-sensitive value estimation and stabilize policy optimization during distributional transition. In addition, CL auxiliary mechanisms are incorporated to mitigate catastrophic forgetting in sequential tasks.

The main contributions of this work are summarized as follows:

\begin{enumerate}
    \item An explainable continual reinforcement learning framework for robotic navigation is proposed, which jointly considers continual adaptation and explainability-oriented policy learning in sequential navigation tasks.

    \item A two-stage behavior cloning mechanism is proposed to balance early-stage knowledge guidance and late-stage autonomous behavior formation, enabling efficient policy learning while preserving the emergence of explainable behavior patterns.

    \item A fuzzy critic is proposed to stabilize the transition from cloning-guided learning to autonomous interaction learning by reducing extrapolation-sensitive value estimation during policy updates.

    \item Experiments are conducted in incremental navigation scenarios, demonstrating that the proposed method effectively combines CL capability with policy explainability.
\end{enumerate}

The rest of this paper is organized as follows. The related work are presented in Section \uppercase\expandafter{\romannumeral2}. Then, the Methodology is discussed in Section \uppercase\expandafter{\romannumeral3}. The experimental setup is discussed in Section \uppercase\expandafter{\romannumeral4}. The results and analysis are described in Section \uppercase\expandafter{\romannumeral5}. Finally, the conclusion is given in Section \uppercase\expandafter{\romannumeral6}.

\section{RELATED WORK}

\subsection{Guided Reinforcement Learning}

Exploration inefficiency remains a major challenge in RL for robotic navigation and control, especially in high-dimensional or sparse-reward environments. To alleviate this issue, many studies have introduced external knowledge or expert experience to guide policy learning and improve sample efficiency.

One representative direction incorporates expert demonstrations or prior policies into RL. For example, expert models, hybrid action selection strategies, and offline-generated trajectories have been employed to constrain the exploration space and accelerate policy initialization~\cite{1,3,4}. In addition, methods such as Hindsight Experience Replay and BC regularization improve learning stability by encouraging policies to reuse historically effective behaviors~\cite{2}. Some studies further reduce learning difficulty through task decomposition and staged policy guidance~\cite{5}.

Recently, fuzzy systems have also been integrated into RL to provide structured and explainable guidance mechanisms. By combining fuzzy-rule reasoning with RL policies, these methods improve exploration stability while maintaining adaptive capability in complex environments~\cite{6,7}.

Inspired by these studies, this research adopts BC to improve policy learning efficiency in the early training stage.

\subsection{Continual Reinforcement Learning}

In long-term robotic navigation, agents are required to continually adapt to changing environments while preserving previously acquired knowledge. Existing CRL methods can generally be divided into three categories: knowledge reuse, experience replay , and parameter regularization.

Knowledge reuse methods improve cross-task adaptation by transferring historical policies or representations. Representative approaches employ policy libraries, attention mechanisms, or persistent policy guidance to improve transfer efficiency and reduce forgetting~\cite{8,9}. Experience replay methods alleviate forgetting by storing and replaying representative experiences, where structured replay and context-aware memory strategies have shown effectiveness in CL scenarios~\cite{10,11,12,CORESETS,SCDMSLL}. Parameter regularization methods preserve historical knowledge by constraining updates to important parameters. Typical approaches include Fisher-information-based EWC \cite{ewc} and trajectory-based SI\cite{si}, together with several extended regularization strategies~\cite{13x}.

Although these approaches have achieved promising results, most existing studies focus on only a single perspective and rarely consider CL together with policy guidance and explainability in a unified framework. Motivated by this observation, the present work integrates knowledge-guided learning, continual regularization, and explainable policy structures into a unified RL framework, thereby enabling both CL capability and policy explainability.

\subsection{Fuzzy Reinforcement Learning}
Due to their explainability and uncertainty modeling capability, fuzzy systems have been widely incorporated into RL to improve control performance and training stability. In robotic control and navigation tasks, fuzzy RL is commonly used for value approximation, policy guidance, and reward design.

One category of methods employs fuzzy systems as function approximators or dynamic models to assist RL optimization. Representative studies utilized fuzzy systems to approximate value functions, model uncertain dynamics, and construct approximately optimal controllers for nonlinear systems~\cite{15,18,20}. These approaches improve modeling capability and system robustness, but fuzzy systems are generally treated only as auxiliary components.

Another category directly integrates fuzzy systems into policy or control architectures. For example, hierarchical frameworks combined RL decision-making with low-level fuzzy controllers~\cite{16}, while several studies fused fuzzy prior policies with RL policies through adaptive weighting mechanisms~\cite{17}. Fuzzy systems have also been employed for reward shaping and auxiliary evaluation in complex control problems~\cite{19}. Although these methods improve explainability and stability, most adopt loosely coupled designs between fuzzy systems and RL policies.

Overall, existing fuzzy RL methods mainly utilize fuzzy systems for auxiliary modeling or external guidance, while deeper integration with policy learning and CL mechanisms remains limited. Motivated by this observation, the proposed framework incorporates fuzzy systems directly into the network structure for value estimation and policy learning, thereby enabling explainability and CL capability within a unified framework.

\subsection{Hierarchical Explainability in Reinforcement Learning}

As RL has been increasingly applied to robotic control and autonomous navigation, policy explainability has attracted growing attention. Existing studies mainly improve explainability from two perspectives: post-hoc explanation and structural explainability.

Post-hoc methods analyze trained policies to estimate the contribution of input features to decision-making. Representative approaches employ Shapley-value-based analysis, semantic feature grouping, and attention mechanisms to explain policy behaviors from the input--output perspective~\cite{21,24,25}. These methods provide intuitive explanation of policy decisions without modifying the original learning framework.

Another line of research improves explainability through policy structure design. For example, some studies modeled policies using graph-based structures to explicitly represent decision dependencies~\cite{22}, while others introduced intermediate semantic variables or auxiliary tasks to generate explainable latent representations during decision-making~\cite{23}. Compared with post-hoc analysis, these approaches provide more direct structural explanation of the policy process.

Overall, existing explainable RL methods have demonstrated the effectiveness of both external explanation and structural modeling. Motivated by these studies, the present work adopts an intermediate-variable-based explainability framework by introducing semantically meaningful latent variables into the policy architecture, thereby enabling structured explanation of navigation behaviors.

\section{{METHODOLOGY}}
\subsection{Overview}
\begin{figure*}[htbp]
 \centering
  \includegraphics[width=0.99\textwidth,]{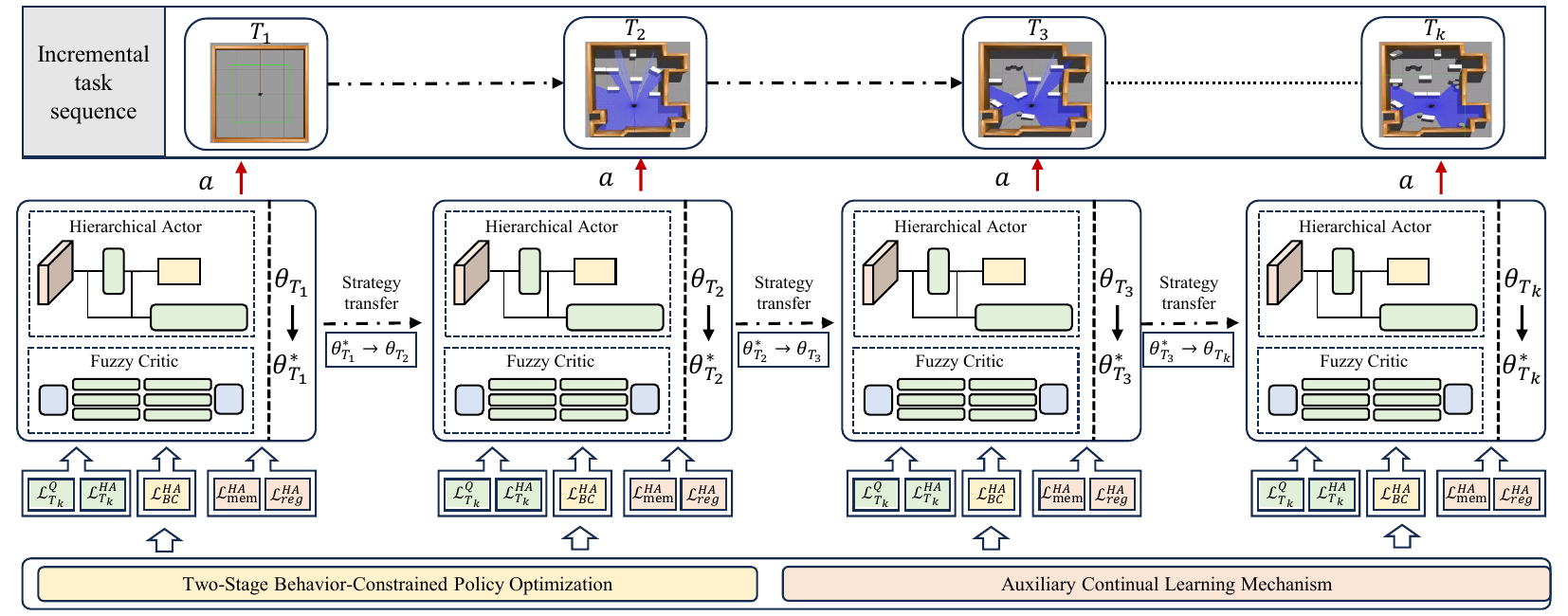}
  \caption{The overall framework of the proposed methodology.}
  \label{fig3.1}
\end{figure*}

The proposed FECHRL framework integrates CRL and explainable policy learning within a unified architecture. The overall framework is illustrated in Fig.\ref{fig3.1}. To preserve previously acquired knowledge during incremental task learning, integrative consolidation and memory augmentation mechanisms are incorporated into the learning framework. To improve policy explainability, a hierarchical actor structure with an intermediate behavior mode layer is introduced, enabling the policy generation process to be represented through semantically meaningful latent modes. Furthermore, a fuzzy critic and a two-stage behavior cloning mechanism are proposed to support stable policy adaptation during CL. The two-stage cloning mechanism balances external policy guidance and autonomous exploration by switching from strong cloning supervision to weak cloning constraints during training. Meanwhile, the fuzzy critic utilizes controllable initialization and activation-based local updates to alleviate critic extrapolation errors during the switching stage, thereby improving learning stability.
\begin{remark}
\textbf{Feature 1: Unified continual and explainable reinforcement learning.}
The proposed framework jointly supports continual task adaptation and semantically explainable policy generation within a unified learning architecture.
\end{remark}

\begin{remark}
\textbf{Feature 2: Two-stage guided policy adaptation.}
The proposed two-stage behavior cloning mechanism balances rapid task adaptation and autonomous policy learning, enabling explainable behavior mode formation during CL.
\end{remark}

\begin{remark}
\textbf{Feature 3: Stable fuzzy critic learning.}
The fuzzy critic improves training stability during the transition from imitation learning to autonomous interaction by reducing critic extrapolation errors.
\end{remark}

\subsection{Problem Formulation}
We consider a CRL setting where an agent sequentially interacts with a series of tasks\(T = [T_1, T_2, \ldots, T_k, \ldots]\). Each task is characterized by distinct environmental dynamics and reward distributions, while sharing the same state and action spaces.

For each task \(T_k\), the agent learns a policy parameterized by \(\theta\) to maximize the expected return:
\begin{equation}
\theta_{T_k}^* = \arg\max_{\theta} J_{T_k}(\theta)
\end{equation}

When a new task \(T_{k+1}\) arrives, the objective is to learn a new policy that maximizes performance on the current task while preserving performance on previously learned tasks. This leads to the following unified optimization problem:
\begin{align}
\theta_{T_{k+1}}^*
=
\arg\max_{\theta}
\quad &
J_{T_{k+1}}(\theta)
+
\lambda \mathcal{U}(\theta)
\\
\text{s.t.} \quad
&
\mathbb{E}[R_{T_i}(\theta)]
\geq
\mathbb{E}[R_{T_i}(\theta_{T_i}^*)] - \varepsilon,
\quad \forall i \leq k
\end{align}

where \(R_{T_i}\) denotes the expected return on task \(T_i\), and \(\varepsilon\) controls the allowable performance degradation across tasks. The term \(\mathcal{U}(\theta)\) represents a general regularization objective that jointly encourages stable CL and structured behavioral organization of the policy, thereby implicitly promoting explainability. The coefficient \(\lambda\) balances task performance and this auxiliary inductive bias.

Due to the sequential and non-revisitable nature of tasks, the above constraints cannot be explicitly enforced during training. Therefore, they are typically approximated using parameter regularization and memory-based replay mechanisms to mitigate catastrophic forgetting while maintaining stable policy learning across tasks.

\subsection{Fuzzy Critic}

To improve the stability and explainability of value estimation in CRL, this study replaces the conventional neural-network critic with a Takagi--Sugeno (TS) fuzzy critic. Traditional actor--critic methods generally approximate the value function using a globally coupled neural network:
\begin{equation}
Q(s,a)=f_{\phi}(s,a)
\label{eq:3.1.1}
\end{equation}

Although neural networks possess strong nonlinear approximation capability, their globally shared parameters may cause value overestimation and parameter overwriting during CL. In the proposed two-stage behavior cloning framework, the first stage mainly relies on supervised imitation learning, where most samples correspond to feasible or high-reward trajectories. Consequently, the critic may produce overestimated values for unseen state--action pairs:
\begin{equation}
Q(s_u,a_u)\gg Q^{*}(s_u,a_u)
\label{eq:3.1.2}
\end{equation}
where \(Q^{*}(s,a)\) denotes the true value function. Since the actor update directly depends on the critic gradient,
\begin{equation}
\nabla_{\theta}J(\theta)
=
\mathbb{E}
\left[
\nabla_a Q(s,a)\nabla_{\theta}\pi_{\theta}(s)
\right]
\label{eq:3.1.3}
\end{equation}
incorrect value estimation may further mislead policy optimization during the free exploration stage.

In addition, neural-network critics exhibit globally coupled parameter updates:
\begin{equation}
\phi \leftarrow \phi - \alpha \nabla_{\phi}L_Q
\label{eq:3.1.4}
\end{equation}
where updates induced by new samples may simultaneously affect previously learned regions, resulting in parameter overwriting during CL.

To alleviate these issues, the proposed method employs a TS fuzzy system for value estimation:
\begin{equation}
Q(s,a)=\sum_{k=1}^{K}\bar{w}_k y_k
\label{eq:3.1.5}
\end{equation}
where \(K\) denotes the number of fuzzy rules, \(\bar{w}_k\) is the normalized activation weight, and \(y_k\) is the local consequent output. The \(k\)-th fuzzy rule is defined as:
\begin{equation}
R^k:
\text{IF } x_1 \text{ is } A_1^k
\text{ AND } \cdots
\text{ AND } x_n \text{ is } A_n^k
\text{ THEN } y^k
\label{eq:3.1.6}
\end{equation}
Gaussian membership functions are adopted:
\begin{equation}
\mu_{A_i^k}(x_i)
=
\exp
\left(
-
\frac{(x_i-c_i^k)^2}
{2(\sigma_i^k)^2}
\right)
\label{eq:3.1.7}
\end{equation}
and the normalized rule activation is computed as (\ref{eq:3.1.8})
\begin{equation}
\bar{w}_k
=
\frac{\prod_{i=1}^{n}\mu_{A_i^k}(x_i)}
{\sum_{j=1}^{K}\prod_{i=1}^{n}\mu_{A_i^j}(x_i)}
\label{eq:3.1.8}
\end{equation}

Unlike globally coupled neural-network critics, the TS fuzzy critic possesses localized update characteristics. The consequent parameter gradient satisfies:
\begin{equation}
\frac{\partial Q}{\partial y^k}
=
\bar{w}_k
\label{eq:3.1.9}
\end{equation}
which indicates that weakly activated rules receive only minor parameter updates. Therefore, newly observed samples mainly affect locally activated regions, reducing interference with previously learned knowledge.

In addition, the consequent outputs are initialized with small values in this work, allowing unseen regions to maintain conservative value estimation during early exploration. Consequently, the proposed fuzzy critic alleviates Q-value overestimation and parameter overwriting during CL, thereby improving the stability of policy optimization and supporting smooth transition in the proposed two-stage behavior cloning framework.

\subsection{Hierarchical Actor Architecture}

This study introduced a hierarchical actor architecture with intermediate behavioral patterns to establish a semantic association between high-level environmental perception and low-level motion control.

First, high-dimensional sensory observations obtained from LiDAR and other perception sensors were mapped into low-dimensional latent representations through a convolutional encoding network:
\begin{equation}
z=f_{\mathrm{enc}}(s_{\mathrm{lidar}})
\label{eq:3.2.1}
\end{equation}
where \(f_{\mathrm{enc}}(\cdot)\) denotes the convolutional feature extraction network with parameters $W$, and \(z\) represents the compact latent representation of the environmental state.

Subsequently, a behavioral pattern generation layer was introduced to further project the latent features into a low-dimensional behavioral pattern space:
\begin{equation}
m=g_{\psi}(z)
\label{eq:3.2.2}
\end{equation}
where \(m\) denotes the high-level behavioral tendency of the agent under the current state.

Finally, the low-level policy network generated the control action by jointly considering the latent state representation, behavioral pattern, and robot state information:
\begin{equation}
a=\pi_{\theta}(z,m,s_{\mathrm{bot}},s_g)
\label{eq:3.2.3}
\end{equation}
where \(\pi_{\theta}(\cdot)\) denotes the low-level action policy.

By introducing the behavioral pattern variable, the actor network was able to preserve semantic behavioral information during the action generation process, thereby improving the explainability of policy execution and enhancing the structural decoupling between high-level decision tendencies and low-level control actions.

\subsection{Two-Stage Behavior-Constrained Policy Optimization}

BC can effectively improve the early-stage learning efficiency of RL by introducing external demonstrations:
\begin{equation}
L_{BC}
=
\mathbb{E}_{(s,a^{*})\sim D_d}
\left[
\left\|
\pi_{\theta}(s)-a_{ks}
\right\|^2
\right]
\label{eq:3.3.1}
\end{equation}
where \(D_d\) denotes the demonstration dataset and \(a_{ks}\) is the demonstrated action.

However, maintaining strong cloning constraints throughout the entire training process may suppress autonomous exploration and restrict the evolution of the intermediate behavior pattern variable \(m\), resulting in reduced behavioral diversity and explainability. To balance learning efficiency and autonomous adaptation, a Two-Stage Behavior-Constrained Policy Optimization strategy is proposed.

In the first stage, a large cloning weight is employed to rapidly establish fundamental navigation behaviors:
\begin{equation}
L_{\mathrm{HA}}
=
L_{RL}
+
\lambda_h L_{BC}
+
L_{reg}
\label{eq:3.3.2}
\end{equation}
where \(L_{RL}\) denotes the RL objective and \(L_{reg}\) represents CL regularization terms.

After the policy acquires basic navigation capability, the cloning constraint is weakened to release exploration ability:
\begin{equation}
L_{\mathrm{HA}}
=
L_{RL}
+
\lambda_l L_{BC}
+
L_{reg},
\qquad
\lambda_l \ll \lambda_h
\label{eq:3.3.3}
\end{equation}

Assuming the total number of training episodes is \(T_T\), the switching point is defined as (\ref{eq:3.3.4}).
\begin{equation}
T_s=\eta T_T
\label{eq:3.3.4}
\end{equation}
where \(\eta \in (0,1)\) is the stage ratio. A hard switching strategy is adopted:
\begin{equation}
\lambda_{BC}(t)=
\begin{cases}
\lambda_h, & t<T_s \\
\lambda_l, & t\geq T_s
\end{cases}
\label{eq:3.3.5}
\end{equation}

In this work, the first \(20\%\) of training episodes employ strong cloning constraints with \(\lambda_h=10\), while the remaining stage uses weak constraints with \(\lambda_l=0.01\). The final optimization objective is:
\begin{equation}
L_{\mathrm{HA}}=
\begin{cases}
L_{RL}+L_{BC}+L_{reg}, & t<0.2T \\
L_{RL}+0.01L_{BC}+L_{reg}, & t\geq0.2T
\end{cases}
\label{eq:3.3.6}
\end{equation}

Besides hard switching, the proposed framework can also be extended to progressive or soft switching strategies:
\begin{equation}
\lambda_{BC}(t)=\lambda_h e^{-kt}
\label{eq:3.3.7}
\end{equation}
\begin{equation}
\lambda_{BC}(t)
=
\lambda_l
+
\frac{\lambda_h-\lambda_l}
{1+\exp\left(\beta(t-T_s)\right)}
\label{eq:3.3.8}
\end{equation}
where \(k\) and \(\beta\) denote decay coefficients.

\subsection{Auxiliary Continual Learning Mechanism}

Since this work was conducted under a CRL setting, the policy network was prone to catastrophic forgetting during incremental task learning. To improve knowledge retention capability, this study adopted the stabilization mechanism proposed in~\cite{5X}, which combines parameter consolidation and memory replay strategies. As this component was not the primary focus of this work, only the key procedures are briefly introduced.

\subsubsection{Parameter Consolidation}

To reduce the destructive influence of new task learning on previously acquired knowledge, an integrated parameter importance estimation strategy was adopted:
\begin{equation}
\Omega_i^k
=
\alpha \Upsilon_i^k
+
(1-\alpha)
F_{T_k,i}^{HA}
\label{eq:3.4.1}
\end{equation}
where
\begin{equation}
\Upsilon_i^k
=
\frac{I_i^k}
{|I_i^k|+\epsilon},
\qquad
F_{T_k,i}^{HA}
=
\mathbb{E}_{s\sim D_k}
\left[
\left(
\frac{\partial HA(s|\theta)}
{\partial \theta_i}
\right)^2
\right]
\label{eq:3.4.2}
\end{equation}
represent the trajectory-contribution importance and Fisher information importance, respectively. $HA$ represents Hierarchical Actor. 

Based on the estimated importance, the regularization constraint was defined as (\ref{eq:3.4.3}).
\begin{equation}
L_{ic}
=
c
\sum_i
\Omega_i^k
(\theta_i-\theta_{T_k,i})^2
\label{eq:3.4.3}
\end{equation}
where \(c\) denotes the regularization coefficient. This mechanism constrains updates of important parameters during new task learning.

\subsubsection{Memory Replay}

Besides parameter consolidation, a dual-buffer replay mechanism was introduced to preserve representative historical data. Current-task samples were stored in a prioritized replay buffer, where the sampling priority was determined according to the TD error:
\begin{equation}
p_i
=
|Q(s_t,a_t)-y_t|
+\epsilon
\label{eq:3.4.4}
\end{equation}

Meanwhile, successful trajectories from previous tasks were stored in a historical memory buffer, where DBSCAN clustering was employed to reduce redundant samples while preserving representative experiences.

During training, historical samples were replayed together with current-task samples, leading to the replay constraint:
\begin{equation}
L_{mem}
=
\frac{1}{|B_{mem}|}
\sum
\|
HA(s)-a
\|^2
\label{eq:3.4.5}
\end{equation}

Accordingly, the final actor optimization objective jointly consisted of the reinforcement learning loss, behavior cloning loss, parameter consolidation term, and replay regularization:
\begin{equation}
L_{\mathrm{HA}}
=
L_{RL}
+
\lambda_{BC}L_{BC}
+
L_{reg}
\label{eq:3.4.6}
\end{equation}
where $L_{reg}=L_{ic}+L_{mem}$.
Through the joint use of parameter consolidation and memory replay, the proposed framework improved policy stability and enhanced historical knowledge retention during CL.

\subsection{Bi-level Optimization Mechanism for Behavioral Pattern Stability}

Since the behavioral pattern variable \(m\) is used for dynamic reward modulation, fully end-to-end joint optimization may lead to unintended reward exploitation, where the pattern generation layer learns to increase returns by implicitly weakening penalty-related reward components rather than forming semantically meaningful behaviors. This can degrade both explainability and generalization. To address this issue, a bi-level optimization mechanism is introduced to decouple behavioral pattern learning from low-level policy optimization.

The overall dynamic reward is defined as (\ref{eq:3.5.1}).
\begin{equation}
r_{\mathrm{dyn}}(s,a,m)
=
r_{\mathrm{task}}(s,a)
+
\sum_{i=1}^{M}
h_{ai}(m)\, r_k(s,a)
\label{eq:3.5.1}
\end{equation}
where \(h_{ai}(m)\) denotes the pattern-dependent reward weighting.

\subsubsection{Inner-Level Policy Optimization}

In the inner loop, the behavioral pattern generator parameter \(\psi\) is fixed, and only the encoder \(W\), actor \(\theta\), and fuzzy critic \(\eta\) are updated. The critic is trained using the Bellman error:
\begin{equation}
\begin{aligned}
L_{Q_{\eta}}
&=
\mathbb{E}
\left[
\left(
Q_{\eta}(s_t,a_t)-y_t
\right)^2
\right],
\\
y_t
&=
r_{\mathrm{dyn}}
+
\gamma Q_{\eta^-}(s_{t+1},\pi_{\theta^-}(s_{t+1}))
\end{aligned}
\label{eq:3.5.2}
\end{equation}

The actor objective follows:
\begin{equation}
\begin{aligned}
J_{\pi}=\mathbb{E}[Q_{\eta}(s,a)]+
\lambda_{BC}L_{BC}
+
L_{reg}
\end{aligned}
\label{eq:3.5.3}
\end{equation}

Since \(m\) is embedded in policy generation, the encoder gradient contains an additional modulation path:
\begin{equation}
\nabla_W J
=
\mathbb{E}
\left[
\nabla_a Q_{\eta}
\left(
\frac{\partial a}{\partial z}
+
\frac{\partial a}{\partial m}
\frac{\partial m}{\partial z}
\right)
\frac{\partial z}{\partial W}
\right]
\label{eq:3.5.4}
\end{equation}

During this stage, the pattern generator is not updated.

\subsubsection{Outer-Level Behavioral Pattern Optimization}

In the outer loop, the encoder and actor are frozen:
\begin{equation}
\nabla_W=0,\quad \nabla_{\theta}=0
\label{eq:3.5.6}
\end{equation}
and only the behavioral pattern generator \(\psi\) is optimized.

To avoid bias introduced by dynamic reward shaping, the outer critic is trained using only the task reward:
\begin{equation}
\begin{aligned}
L_{Q_{\xi}}
&=
\mathbb{E}
\left[
\left(
Q_{\xi}(s_t,a_t)-y_t^{\xi}
\right)^2
\right],
\\
y_t^{\xi}
&=
r_{\mathrm{task}}
+
\gamma Q_{\xi^-}(s_{t+1},\pi_{\theta}(s_{t+1}))
\end{aligned}
\label{eq:3.5.7}
\end{equation}

The gradient for pattern learning is given by:
\begin{equation}
\nabla_{\psi}J_{\mathrm{outer}}
=
\mathbb{E}
\left[
\nabla_a Q_{\xi}(s,a)
\frac{\partial a}{\partial m}
\frac{\partial m}{\partial \psi}
\right]
\label{eq:3.5.8}
\end{equation}

This bi-level formulation ensures that behavioral patterns are optimized only through improvements in true task performance, rather than through reward manipulation, thereby improving stability and preserving explainability. The overall algorithm flow is shown in Algorithm.\ref{alg:bilevel_ac}.

\begin{algorithm}[t]
\caption{FECHRL}
\label{alg:bilevel_ac}
\begin{algorithmic}[1]
\REQUIRE 
$\pi_{\theta}$;
critic $Q_{\eta}$;
critic $Q_{\xi}$;
feature encoder $f_{\mathrm{conv}}(\cdot; W)$;
mode generation network $g_{\psi}(\cdot)$;
fuzzy reward module $F_{\phi}(\cdot)$.

\STATE Initialize $\theta, {\eta}, W, \psi, \phi, \xi$
\STATE Initialize target networks $\theta', {\eta}', \xi'$
\STATE Initialize replay buffer $\mathcal{D}$
\FOR{Task=1,2,...K} 
    \FOR{episode $=1$ to \(T_T\)}
        \STATE Observe initial state $s_0$
        
        \FOR{t $=1$ to $T$}
            \STATE Extract feature $z_t$, build pattern $m_t$, get reward-weight $ha_t$ via fuzzy inference
            \STATE Select action $a_t = \pi_{\theta}(z_t, m_t)$
            \STATE Execute $a_t$, observe $r_t$, $r_t^{\mathrm{task}}$, and $s_{t+1}$
            \STATE Store transition $(s_t, a_t, r_t, r_t^{\mathrm{task}}, s_{t+1},a_{ks})$ in $\mathcal{D}_{Tk}$
        \ENDFOR
    
        \STATE Sample a minibatch data from $\mathcal{D}_{Tk}$ and $\mathcal{D}_{mem}$
    
        \COMMENT{Inner-Level Update (Freeze $\psi$)} 
        \STATE Update  $Q_{\eta}$ using dynamically weighted reward (\ref{eq:3.5.2})
        \STATE Update $\pi_{\theta}$ and encoder $f_{\mathrm{conv}}(\cdot; W)$ via policy gradient (\ref{eq:3.5.3}-\ref{eq:3.5.4})
        \STATE Soft-update target networks $\theta', \eta'$
    
        \COMMENT{Outer-Level Update (Freeze $\theta, W, \eta$)}
        \STATE Update critic $Q_{\xi}$ using task-level reward $r^{\mathrm{task}}$. (\ref{eq:3.5.7})
        \STATE Update mode generation network $g_{\psi}$ by (\ref{eq:3.5.8})
        \STATE Soft-update target network $\xi'$
    \ENDFOR
\ENDFOR
\end{algorithmic}
\end{algorithm}

\section{Experimental Setup}
This study proposed a CRL framework that integrates a fuzzy critic, two-stage behavior cloning, and multi-dimensional parameter regularization, with the objective of jointly improving CL capability and policy explainability in incremental robotic navigation scenarios. Accordingly, the experiments were designed with two primary objectives:

\begin{enumerate}
    \item To verify whether the proposed method can effectively integrate CL capability and policy explainability;
    \item To evaluate and analyze the characteristics of the fuzzy critic and the two-stage behavior cloning mechanism.
\end{enumerate}

This section provides detailed descriptions of the experimental settings.

\subsubsection{Baseline}
To evaluate the effectiveness of the proposed framework in continual policy learning and behavior-aware optimization, several representative RL and CL methods were selected for comparison. All baseline methods were implemented under the same network architecture, training environment, and replay settings to ensure a fair evaluation.

\begin{itemize}
    \item \textbf{DDPG}~\cite{DDPG}:  
    A classical actor--critic RL algorithm for continuous control tasks, which was employed as the fundamental baseline method.

    \item \textbf{EWC-DDPG}~\cite{ewc}:  
    A CRL method that constrains parameter updates using Fisher information regularization.

    \item \textbf{SI-DDPG}~\cite{si}:  
    A CRL approach based on synaptic importance accumulated along the optimization trajectory.

    \item \textbf{CORESET}~\cite{CORESETS}:  
    A replay-based CRL method.

    \item \textbf{SCOMMER}~\cite{SCDMSLL}:  
    A dual-memory CL framework.
\end{itemize}

\subsubsection{Evaluation Metrics}

To evaluate continual policy learning performance, the standard CL metrics of forgetting and forward transfer are adopted.

\begin{itemize}

\item \textbf{Forgetting}:  
This metric measures the performance degradation on previously learned tasks after learning new tasks. It is defined as (\ref{eq4.3.1}).

\begin{equation}\label{eq4.3.1}
F^{(T_k)}
=
\frac{1}{k-1}
\sum_{i=1}^{k-1}
\left(
A_i(T_i)-A_i(T_k)
\right)
\end{equation}

where \(A_i(T_i)\) denotes the performance on task \(T_i\) immediately after training, and \(A_i(T_k)\) denotes the performance on the same task after learning task \(T_k\).

\item \textbf{Forward Transfer}:  
Forward transfer evaluates how previously acquired knowledge facilitates learning subsequent tasks:

\begin{equation}\label{eq4.3.2}
FT^{(T_k)}
=
\frac{1}{k-1}
\sum_{i=1}^{k-1}
A_k(T_i)
\end{equation}

\end{itemize}

\subsection{Training Scenarios and Experimental Environments}
\begin{figure}[ht]
 \centering
 \subfigure[Env1 ]{\includegraphics[width=0.15\linewidth,height=0.07\textheight ]
{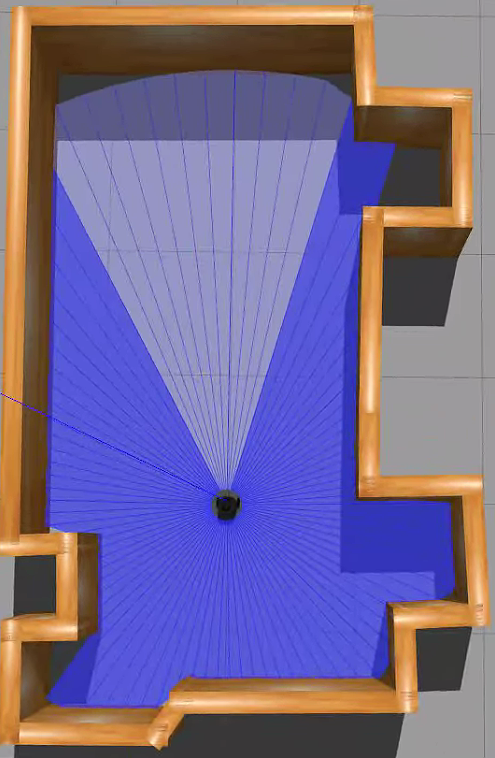}\label{fig: sub_figure4.2.a}}
\subfigure[Env2]{\includegraphics[width=0.15\linewidth, height=0.07\textheight]
{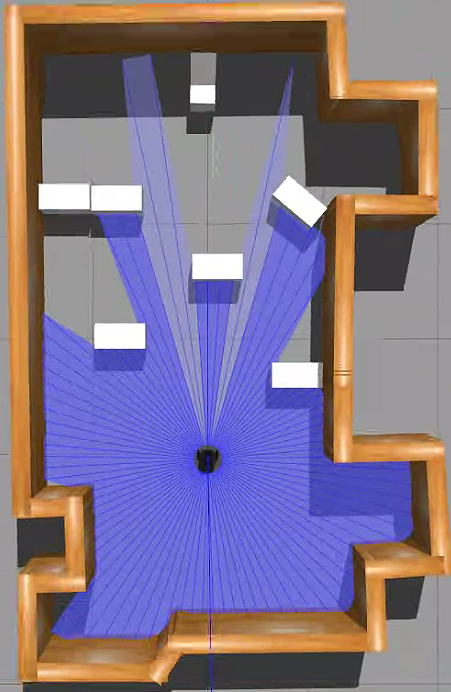}\label{fig: sub_figure4.2.b}}
\subfigure[Env3 ]{\includegraphics[width=0.15\linewidth, height=0.07\textheight]
{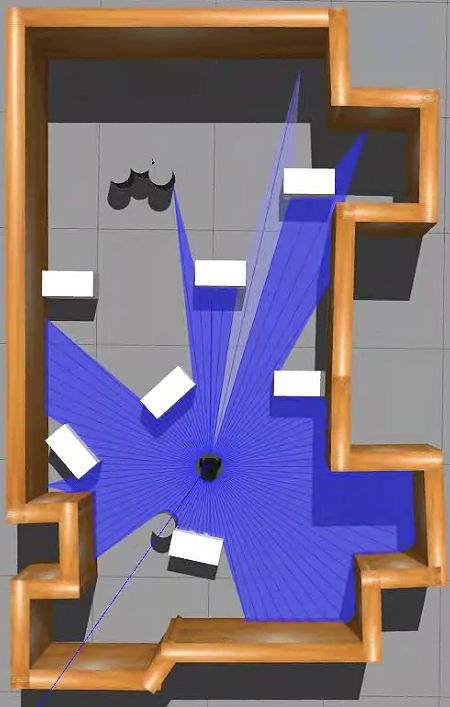}\label{fig: sub_figure4.2.c}}
\subfigure[{Env4}]{\includegraphics[width=0.15\linewidth, height=0.07\textheight]
{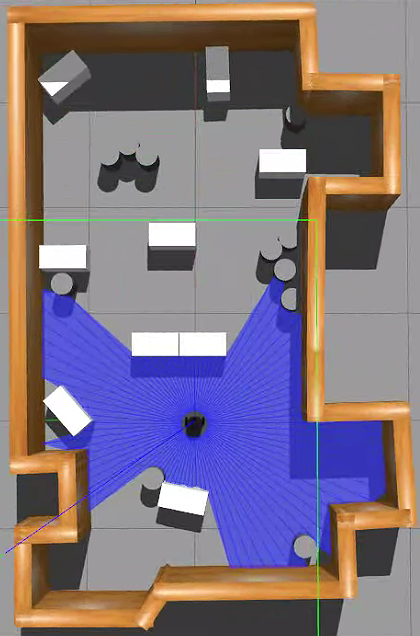}\label{fig: sub_figure4.2.d}}
\subfigure[{Env5}]{\includegraphics[width=0.15\linewidth, height=0.07\textheight]
{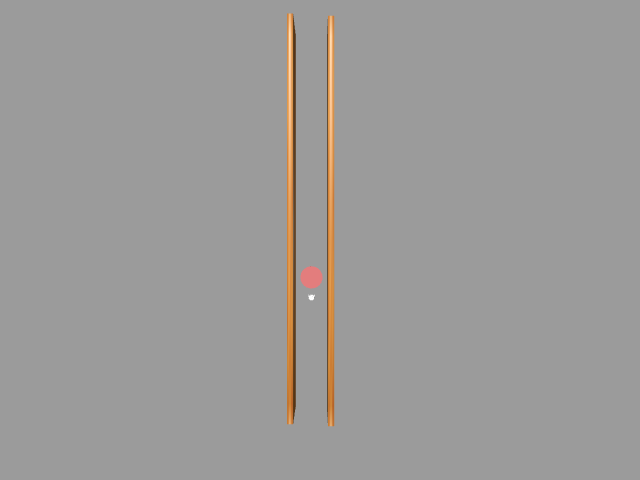}\label{fig: sub_figure4.2.e}}
  \caption{Simulation environment}
  \label{fig4.2}
\end{figure}

Five different simulation environments are constructed to evaluate the performance of FECHRL (see Fig.~\ref{fig4.2}).  
In Env1--Env4, the number of obstacles gradually increases and the layouts continuously change, thereby progressively increasing the environmental complexity. These environments are mainly used to evaluate continual learning capability.  
Env5 is a narrow corridor scenario in which the optimal behavior mode is relatively easy to identify, and is therefore used to evaluate policy explainability.

\begin{figure}[ht]
 \centering
 \subfigure[Env1 ]{\includegraphics[width=0.45\linewidth,height=0.25\linewidth]
{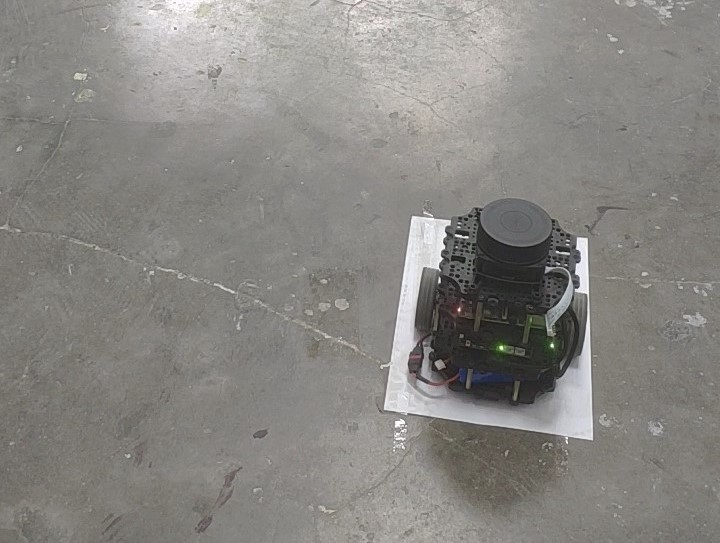}\label{fig: sub_figure4.2.1.a}}
\subfigure[Env2 ]{\includegraphics[width=0.45\linewidth, ]
{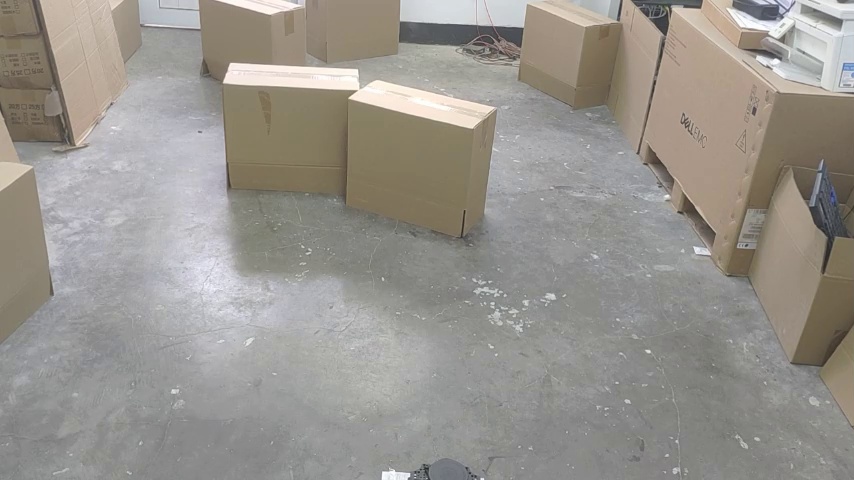}\label{fig: sub_figure4.2.1.b}}
  \caption{Real environment}
  \label{fig4.2.1}
\end{figure}

To verify the ability of simulation to reality, two scenarios (see Fig.\ref{fig4.2.1}) were constructed with different obstacle distributions and sizes from the training scenario.

\subsubsection{Network Architecture}
\begin{figure*}[htbp]
 \centering
\includegraphics[width=0.95\textwidth]{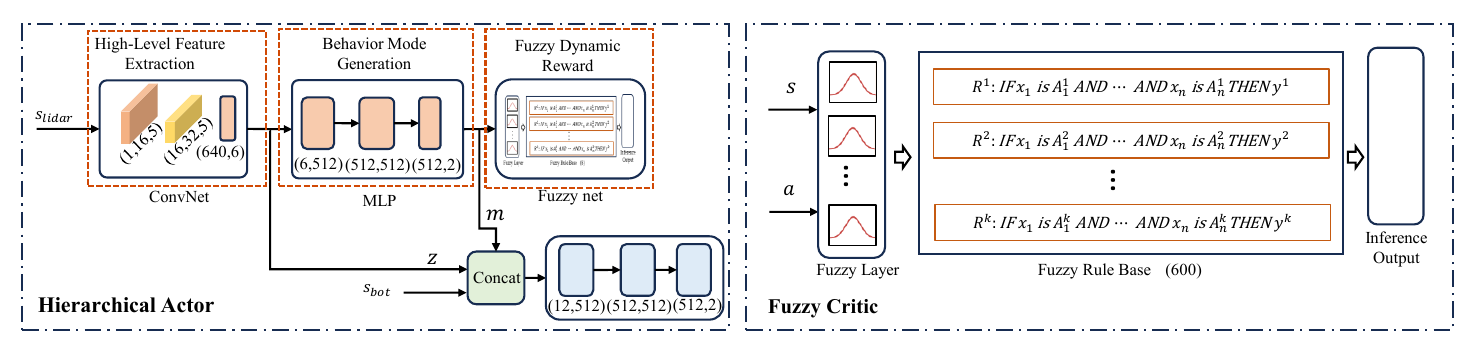}
  \caption{Implementation framework}
  \label{fig4.3}
\end{figure*}

The overall network architecture is illustrated in Fig.~\ref{fig4.3}.  
The actor network consists of three layers. The feature extraction layer receives 90-dimensional point cloud inputs and performs feature compression. The intermediate layer generates behavior modes, which are concatenated with the 2-dimensional target state and 2-dimensional action command as the input to the low-level policy network.  The fuzzy critic receives the concatenated state-action vector and performs inference using 600 fuzzy rules. In addition, the fuzzy dynamic reward network is implemented as a fixed-parameter network to generate dynamic reward weights.  Table~\ref{tab4.1} summarizes the detailed hyperparameter settings.
\begin{table}[H]
\scriptsize
 \caption{PARAMETER SETTINGS}
 \label{tab4.1}
 \renewcommand\arraystretch{1.1}
 \begin{center}
 \setlength{\tabcolsep}{5mm}{
 \begin{tabular}{cc}
 \hline
Items  & Value \\ \hline
optimizer $\pi$ / $g_\psi$ / $Q_\eta$ / $Q_\xi$  & Adam  lr:0.0003/0.0003/0.001/0.01  \\
$\gamma$ / $\lambda$ / $\tau$ $c$ & 0.99/1/0.05/ 0.1 /0.1\\
Buffer\_size/$\mathcal{B}_\text{task}$/$\mathcal{B}_\text{mem}$ & 100000/256/64\\
$Episodes$ & 300(Env1) 1000(Env2-Env4) 500(Env5)\\
max\_steps & 200(Env1),350(Env2-Env4),500(Env5)\\
 \hline
\end{tabular}
 }
 \end{center}
\end{table}


The reward function consists of a task-level objective and several behavior-aware auxiliary terms:

\begin{equation}\label{eq4.3.5}
\begin{aligned}
r
=
r_{\text{task}}
&- ha_1 \big(10(0.2-v)\big)^2
- ha_2 \omega^2
\\
&+ ha_3 \left(\frac{2d_0}{d_0+d}-1\right)
- ha_4 |\theta_g|
\\
&- 20\,ha_5\,\mathbf{I}(d_{\min}<0.22)
\end{aligned}
\end{equation}

where \(v\) and \(\omega\) denote the linear and angular velocities, \(d_0\) and \(d\) are the initial and current goal distances, \(\theta_g\) is the goal heading error, and \(d_{\min}\) denotes the minimum obstacle distance.

The task-level reward is defined as (\ref{eq4.3.5x}).
\begin{equation}\label{eq4.3.5x}
\begin{aligned}
r_{\text{task}}
& =
2500\,\mathbf{I}(\text{reached})
-
2000\,\mathbf{I}(\text{collision or timeout})
\\
&
+
\frac{1000}{|P_L+0.15-G_d|+0.001}
+
\frac{1000}{|E_T-G_T|+0.001}
\end{aligned}
\end{equation}

where \(P_L\) is the path length, \(G_d\) denotes the goal distance, \(E_T\) is the episode duration, and \(G_T\) represents the minimum expected arrival time.

\section{RESULTS AND ANALYSIS}

\subsection{Incremental Task Training Comparison}
As environmental complexity gradually increases, the training process is divided into four sequential navigation tasks. \(T_1\) corresponds to a simple open environment and is trained for 300 episodes, while the more challenging tasks \(T_2\)-\(T_4\) are each trained for 1000 episodes. The success rate is adopted as the primary evaluation metric.

\begin{figure}[htbp]
 \centering
  \includegraphics[width=0.48\textwidth]{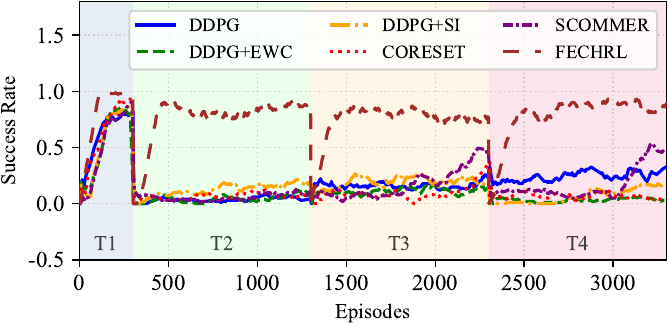}
  \caption{Training success rates of different RL methods}
  \label{fig5.1}
\end{figure}

Fig.~\ref{fig5.1} illustrates the training success rates of different RL methods, including DDPG, DDPG+EWC, DDPG+SI, CORESET, SCOMMER, and the proposed FECHRL framework. For fairness, all baseline methods follow the same experimental settings and training protocols as our previous continual navigation study~\cite{5X}.

Overall, all methods achieve relatively high success rates on the simple \(T_1\). However, as task complexity increases, most baseline methods exhibit degraded adaptability in subsequent tasks. Consolidation-based methods can alleviate catastrophic forgetting to some extent, but the imposed parameter constraints reduce policy adaptability under new environmental distributions. In contrast, SCOMMER demonstrates performance improvements during the later stages of \(T_3\) and \(T_4\), indicating that its dual-memory mechanism provides relatively stable knowledge transfer during CL.

The proposed FECHRL framework maintains stable learning performance across all incremental tasks and rapidly recovers policy performance after task transitions. This suggests that the introduced guided learning mechanism effectively improves policy learning efficiency during the early stages of new tasks while reducing performance degradation caused by unstable exploration during task switching.

\subsection{Quantitative Analysis of Sequential Task Performance}
To further evaluate the CL capability of the proposed framework, quantitative experiments were conducted over four sequential navigation tasks. The baseline CL methods follow the same evaluation protocol and experimental settings adopted in our previous continual navigation study~\cite{5X}. The results averaged over five random seeds are reported in Table~\ref{tab5.1}.

Table~\ref{tab5.1} presents the performance of different methods after sequentially learning tasks \(T_1\)-\(T_4\). Here, \(MT_k\) denotes the model obtained after completing the training of task \(T_k\). The last two columns report the average forgetting rate \(F^{T_k}\) and forward transfer rate \(FT^{T_k}\), respectively.

Overall, the baseline methods exhibit different degrees of catastrophic forgetting during sequential learning. In particular, KGDDPG shows severe performance degradation on previously learned tasks after training subsequent environments, indicating limited knowledge retention capability. Although KGEWC and KGSI alleviate forgetting through parameter consolidation, their performance on earlier tasks still decreases significantly after later-stage training, suggesting that consolidation alone cannot fully balance stability and adaptability in continual navigation scenarios.

In contrast, the proposed FECHRL framework demonstrates strong CL capability across all sequential tasks. As shown in Table~\ref{tab5.1}, FECHRL maintains consistently high success rates on previously learned tasks even after completing all subsequent training stages. For example, the success rate on \(T_2\) further improves from \(0.88\) at \(MT_2\) to \(0.94\) at \(MT_4\), while the performance on \(T_1\) remains close to perfect throughout the entire learning process. Moreover, the forgetting metrics of FECHRL remain negative across all stages, indicating that subsequent task learning not only alleviates catastrophic forgetting but also promotes knowledge refinement on previous tasks.

FECHRL also achieves competitive forward transfer performance. In particular, the framework obtains forward transfer rates of \(0.221\) and \(0.200\) on later tasks, demonstrating that the learned behavioral representations and guided policy adaptation mechanism provide effective prior knowledge for unseen environments. These results indicate that the proposed framework is capable of simultaneously preserving historical knowledge and supporting efficient adaptation to new tasks, thereby achieving stable CRL performance.

\renewcommand{\arraystretch}{1.0}
\begin{table*}[htbp]
\caption{Performance after training each task across different methods.}
 \label{tab5.1}
\centering
\begin{tabular}{lccccccc}
\toprule
\textbf{Method}&      & \multicolumn{4}{c}{\textbf{Train after}}  \\
              &      & MT1  & MT2  & MT3  & MT4  & $F^{T_k}$ & $FT^{T_k}$        \\
\midrule
\multirow{4}{*}{KGDDPG} 
& T1 & 0.91$\pm 0.01$ & 0.91$\pm 0.01$ & 0.90$\pm 0.00$ & 0.75$\pm 0.01$ & -     & -      \\
& T2 & 0.09$\pm 0.02$ & 0.79$\pm 0.05$ & 0.61$\pm 0.05$ & 0.43$\pm 0.01$ & 0.000$\pm 0.026$ & 0.084$\pm 0.017$   \\
& T3 & 0.01$\pm 0.00$ & 0.46$\pm 0.05$ & 0.78$\pm 0.05$ & 0.42$\pm 0.04$ & 0.110$\pm 0.034$ & 0.162$\pm 0.021$ \\
& T4 & 0.02$\pm 0.01$ & 0.03$\pm 0.02$ & 0.14$\pm 0.01$ & 0.80$\pm 0.07$ & 0.296$\pm 0.040$ & 0.131$\pm 0.127$   \\
\midrule
\multirow{4}{*}{KGEWC} 
& T1 & 0.75$\pm 0.06$ & 0.77$\pm 0.01$ & 0.73$\pm 0.01$ & 0.55$\pm 0.02$ & -     & -      \\
& T2 & 0.05$\pm 0.02$ & 0.91$\pm 0.00$ & 0.70$\pm 0.03$ & 0.50$\pm 0.05$ & -0.002$\pm 0.065$  & 0.050$\pm 0.023$   \\
& T3 & 0.00$\pm 0.00$ & 0.36$\pm 0.02$ & 0.78$\pm 0.02$ & 0.62$\pm 0.03$ & 0.128$\pm 0.044$ & 0.115$\pm 0.013$ \\
& T4 & 0.00$\pm 0.00$ & 0.12$\pm 0.02$ & 0.09$\pm 0.00$ & 0.74$\pm 0.10$ & 0.261$\pm 0.017$ & 0.099$\pm 0.011$   \\
\midrule
\multirow{4}{*}{KGSI} 
& T1 & 0.93$\pm 0.02$ & 0.96$\pm 0.01$ & 0.82$\pm 0.02$ & 0.72$\pm 0.02$ & -     & -      \\
& T2 & 0.06$\pm 0.02$ & 0.87$\pm 0.02$ & 0.53$\pm 0.01$ & 0.56$\pm 0.02$ & -0.024$\pm 0.027$& 0.060$\pm 0.021$   \\
& T3 & 0.00$\pm 0.00$ & 0.51$\pm 0.02$ & 0.81$\pm 0.00$ & 0.71$\pm 0.01$ & 0.227$\pm 0.033$ & 0.158$\pm 0.012$ \\
& T4 & 0.00$\pm 0.00$ & 0.11$\pm 0.01$ & 0.20$\pm 0.01$ & 0.79$\pm 0.07$ & 0.209$\pm 0.029$ & 0.139$\pm 0.006$   \\
\midrule
\multirow{4}{*}{\makecell{FECHRL}} 
& T1 & 0.95$\pm 0.00$ & 1.00$\pm 0.00$ & 1.00$\pm 0.00$  & 1.00$\pm 0.01$  & -     & -      \\
& T2 & 0.01$\pm 0.00$ & 0.88$\pm 0.01$ & 0.89$\pm 0.01$  & 0.94$\pm 0.01$  & -0.048$\pm 0.004$ & 0.012$\pm 0.004$ \\
& T3 & 0.01$\pm 0.00$ & 0.85$\pm 0.02$ & 0.82$\pm 0.03$  & 0.90$\pm 0.01$  & -0.033$\pm 0.009$ & 0.221$\pm 0.006$ \\
& T4 & 0.00$\pm 0.00$ & 0.26$\pm 0.02$ & 0.22$\pm 0.00$  & 0.89$\pm 0.01$  & -0.064$\pm 0.010$ & 0.200$\pm 0.004$ \\
\bottomrule
\end{tabular}
\end{table*}

\subsection{Two-stage Cloning Analysis}
In Section \uppercase\expandafter{\romannumeral3}, three behavior cloning switching strategies, namely Hard switching, Progressive switching, and Soft switching, were introduced. To further analyze the differences among these strategies, comparisons are conducted from two perspectives: training performance, testing results.

\begin{figure}[htbp]
 \centering
  \includegraphics[width=0.48\textwidth,]{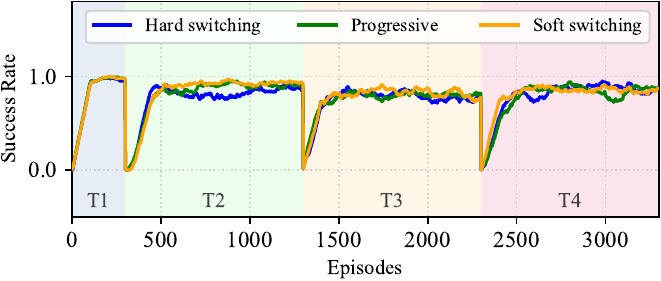}
  \caption{Training success rates of different switching}
  \label{fig5.2}
\end{figure}

The training results of the three switching strategies are shown in Fig.\ref{fig5.2}. Overall, all strategies exhibit similar success-rate trends during CL across multiple tasks. However, Hard switching achieves lower success rates than the other two strategies during the middle stage of learning in \(T_2\), followed by a subsequent recovery in the later stage. This phenomenon is mainly caused by the earlier switching point of Hard switching, which leaves more time for autonomous exploration and interaction. Therefore, the temporary performance decline is associated with increased exploration capability.

Table~\ref{tab5.2} further compares the effects of different BC switching strategies, including Progressive switching and Soft switching. Compared with the Hard switching strategy adopted in FECHRL, both methods maintain relatively large cloning constraints over a longer training period, resulting in different CL characteristics.

As shown in the table, both Progressive switching and Soft switching achieve high performance on previously learned tasks, indicating that stronger cloning supervision can effectively stabilize policy learning. However, compared with FECHRL, these two strategies exhibit weaker adaptability in continual tasks. For example, although Progressive switching achieves relatively high forward transfer on \(T_3\) and \(T_4\), its forgetting metrics gradually become positive, suggesting that excessive dependence on cloning guidance restricts policy plasticity during later-stage adaptation. A similar phenomenon can also be observed in Soft switching, where the delayed decay of the cloning loss preserves behavioral consistency but limits autonomous exploration capability, resulting in reduced performance under unseen environmental distributions.

In contrast, the Hard switching strategy employed in FECHRL enforces an early transition from strong imitation guidance to weak cloning constraints during training. This mechanism rapidly releases policy exploration capability, enabling the agent to learn from environmental interaction more effectively. Consequently, FECHRL achieves lower forgetting rates and stronger forward transfer performance across continual tasks. These results indicate that appropriately reducing imitation constraints during the intermediate stage of training is essential for balancing knowledge guidance and policy adaptability in continual reinforcement learning.

Overall, different switching strategies can be flexibly selected according to different task requirements. Moreover, through appropriate parameter adjustment, these strategies can achieve approximately equivalent learning performance under different CL scenarios.

\begin{table*}[htbp]
\caption{{Mean performance after training each task for  Progressive and  Soft switching.}}
\label{tab5.2}
\centering
{
\begin{tabular}{lccccccc}
\toprule
\textbf{Method}&      & \multicolumn{4}{c}{\textbf{Train after}}  \\
              &      & MT1  & MT2  & MT3  & MT4  & $F^{T_k}$ & $FT^{T_k}$        \\
\midrule
\multirow{4}{*}{Progressive} 
& T1 & 0.97$\pm0.01$ & 1.00$\pm0.00$ & 1.00$\pm0.00$ & 1.00$\pm0.00$ & -     & -      \\
& T2 & 0.03$\pm0.01$ & 0.91$\pm0.01$ & 0.84$\pm0.02$ & 0.83$\pm0.01$ & -0.030$\pm0.008$  & 0.033$\pm0.005$   \\
& T3 & 0.02$\pm0.01$ & 0.76$\pm0.01$ & 0.92$\pm0.01$ & 0.93$\pm0.02$ & 0.020$\pm0.014$ & 0.212$\pm0.005$ \\
& T4 & 0.00$\pm0.00$ & 0.27$\pm0.03$ & 0.33$\pm0.03$ & 0.89$\pm0.02$ & 0.013$\pm0.005$ & 0.208$\pm0.010$   \\
\midrule
\multirow{4}{*}{Soft switching} 
& T1 & 0.97 $\pm0.01$ & 1.00$\pm0.00$ & 1.00$\pm0.00$  & 1.00$\pm0.00$ & -      & -      \\
& T2 & 0.01 $\pm0.00$ & 0.95$\pm0.01$ & 0.86$\pm0.01$  & 0.88$\pm0.01$ & -0.027$\pm0.005$ & 0.010$\pm0.000$    \\
& T3 & 0.00 $\pm0.00$ & 0.64$\pm0.02$ & 0.84$\pm0.05$  & 0.80$\pm0.01$ & 0.028$\pm0.005$ & 0.164$\pm0.004$ \\
& T4 & 0.00 $\pm0.01$ & 0.11$\pm0.01$ & 0.42$\pm0.03$  & 0.83$\pm0.00$ & 0.029$\pm0.016$ & 0.168$\pm0.002$ \\
\bottomrule
\end{tabular}}
\end{table*}

\subsection{Ablation Study}
\subsubsection{Fuzzy Critic Ablation}
\begin{figure}[htbp]
\centering
\includegraphics[width=0.45\textwidth,]{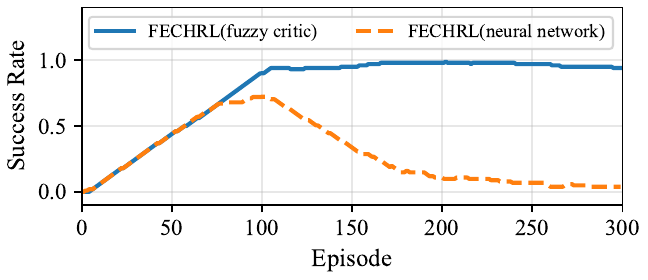}
  \caption{Comparison of different critics}
  \label{fig5.4.1}
\end{figure}

Due to the coupled structure and parameter representation of neural-network-based critics, the values of unseen state--action pairs cannot be explicitly initialized to low values. Consequently, when switching from BC to autonomous interaction, the extrapolation error of the critic may provide incorrect gradient directions, causing the policy initialized by cloning to deviate from the correct optimization trajectory and leading to unstable learning. To verify this issue and demonstrate the effectiveness of the proposed fuzzy critic, we compare the training performance of a neural-network-based critic and the fuzzy critic on task T1. The results are shown in Fig.\ref{fig5.4.1}.
The learning process switches from cloning-based training to autonomous interaction at episode 60. As shown in the figure, under the neural-network critic, the success rate begins to decline after approximately 40 episodes of autonomous learning, indicating that the policy gradually deviates from the stable behavior initialized by cloning. As autonomous interaction continues, this deviation becomes increasingly severe and eventually results in training collapse. In contrast, the fuzzy critic maintains a continuous improvement in success rate after the switching stage, demonstrating that the transition process remains stable. This indicates that the fuzzy critic effectively suppresses extrapolation error and preserves stable value estimation during policy adaptation. Furthermore, the subsequent learning process remains within the normal fluctuation range of stochastic exploration. These results demonstrate the positive role of the fuzzy critic in stabilizing the learning process and validate the rationality of its design within the proposed framework.

\subsubsection{Two-stage-clone Ablation}
\begin{figure}[htbp]
\centering
\includegraphics[width=0.45\textwidth,]{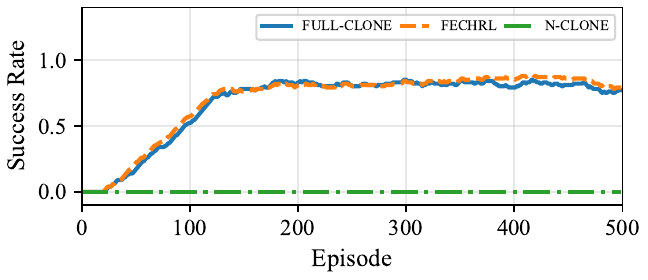}
  \caption{Comparison of different critics}
  \label{fig5.4.2}
\end{figure}

In CL, previously learned policies introduce behavioral biases into new tasks, thereby restricting policy exploration and leading to negative transfer. BC can alleviate this issue by providing external guidance; however, full-stage cloning continuously constrains policy updates and suppresses autonomous exploration as well as behavioral mode evolution. To verify this phenomenon and to investigate the importance of the proposed two-stage cloning mechanism for explainable policy formation, the proposed method is compared with two variants: no cloning (N-CLONE) and full-stage cloning (FULL-CLONE). The corresponding results are shown in Fig.\ref{fig5.4.2}.

As shown in Fig.\ref{fig5.4.2}, in the narrow corridor environment, relying solely on random exploration makes it difficult to obtain effective interaction data, resulting in the failure of N-CLONE. In contrast, both FULL-CLONE and Two-stage-clone achieve significantly improved success rates, demonstrating that behavior cloning effectively facilitates new task learning. From the perspective of continual learning, this process can also be regarded as correcting the behavioral bias induced by previously learned policies. Moreover, FULL-CLONE and Two-stage-clone exhibit similar success-rate trends, indicating that both methods possess strong task-learning capability.

\begin{figure}[ht]
 \centering
 \subfigure[FULL-CLONE ]{\includegraphics[width=0.8\linewidth, ]
    {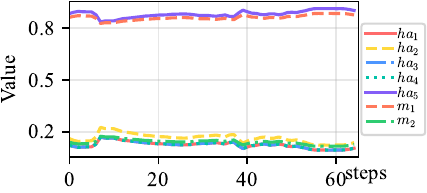}\label{fig: sub_figure5.4.3.a}}
\subfigure[TWO-STAGE]{\includegraphics[width=0.8\linewidth,]
    {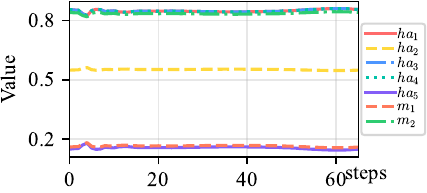}\label{fig: sub_figure5.4.3.b}}
  \caption{Comparison of Behavior Modes under Different Clone Switching}
  \label{fig5.4.3}
\end{figure}

To further analyze the differences in behavioral mode formation, the five sub-reward weights ($ha_i$) and the outputs of the mode layer ($m_i$) under the same target condition are compared, as shown in Fig.\ref{fig5.4.3}. Fig.\ref{fig: sub_figure5.4.3.a} shows that FULL-CLONE forms a behavior mode that consistently favors safety, where the collision-related weight $ha_5$ remains high while the turning-related weight $ha_2$ stays low. This indicates that FULL-CLONE mainly learns the inherent behavioral bias of the cloned policy rather than an environment-adaptive behavior mode. For the narrow corridor environment, a more suitable behavior mode should suppress unnecessary turning while encouraging motion efficiency, meaning that $ha_1$ and $ha_2$ should remain relatively high and the efficiency-related mode $m_2$ should dominate. As shown in Fig.\ref{fig: sub_figure5.4.3.b}, the proposed two-stage cloning strategy gradually learns behavior modes that are more consistent with the environmental characteristics, demonstrating that appropriately releasing autonomous exploration capability is essential for explainable behavioral mode formation.

\subsection{Explainability}
The explainability of the proposed method is reflected in two aspects: behavior mode learning and internal reasoning. Specifically, the learned behavior modes and dynamic reward weights provide semantic explanations for policy actions, while the internal reasoning process of reward generation can be further visualized through rule activations. The objective of this analysis is not only to explain the final navigation actions, but also to verify whether the learned intermediate variables preserve human-understandable semantic meanings.

\begin{figure}[ht]
 \centering
 \subfigure[a]{\includegraphics[width=0.45\linewidth, height=0.36\linewidth]
{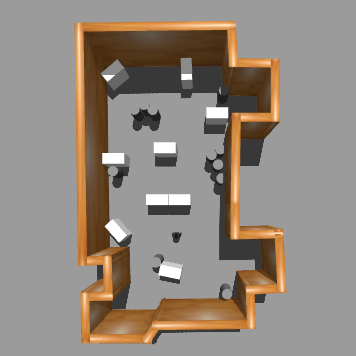}\label{fig: sub_figure5.6.1.a}}
\subfigure[b]{\includegraphics[width=0.45\linewidth, ]
{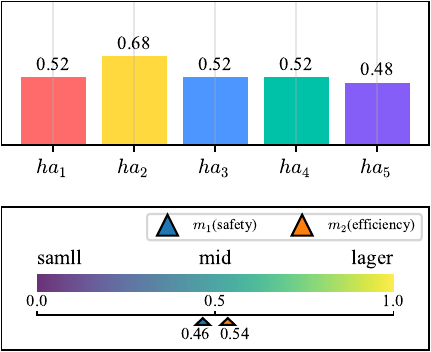}\label{fig: sub_figure5.6.1.b}}
  \caption{Status, reward weight, and mode for steps(0)}
  \label{fig5.6}
\end{figure}

\begin{figure}[ht]
 \centering
 \subfigure[]{\includegraphics[width=0.45\linewidth,height=0.36\linewidth]
{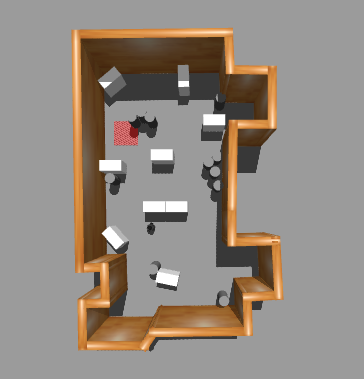}\label{fig: sub_figure5.7.1.a}}
\subfigure[]{\includegraphics[width=0.45\linewidth, ]
{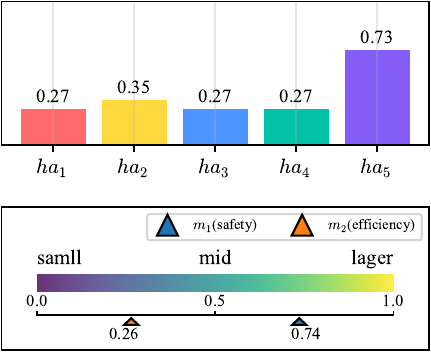}\label{fig: sub_figure5.7.1.b}}
  \caption{Status, reward weight, and mode for steps(22)}
  \label{fig5.7}
\end{figure}

\begin{figure}[ht]
 \centering
 \subfigure[]{\includegraphics[width=0.45\linewidth, height=0.36\linewidth]
{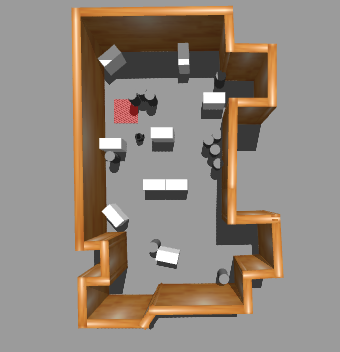}\label{fig: sub_figure5.8.1.a}}
\subfigure[]{\includegraphics[width=0.45\linewidth, ]
{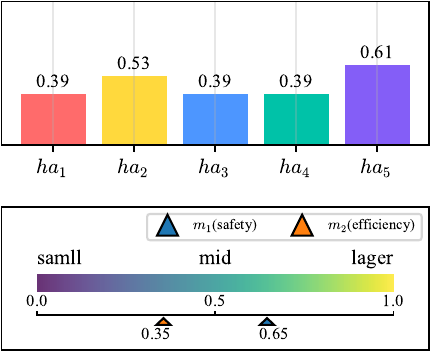}\label{fig: sub_figure5.8.1.b}}
  \caption{Status, reward weight, and mode for steps(61)}
  \label{fig5.8}
\end{figure}

\begin{figure}[ht]
 \centering
 \subfigure[]{\includegraphics[width=0.45\linewidth,height=0.36\linewidth ]
{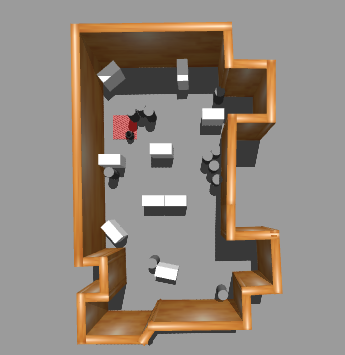}\label{fig: sub_figure5.9.1.a}}
\subfigure[]{\includegraphics[width=0.45\linewidth, ]
{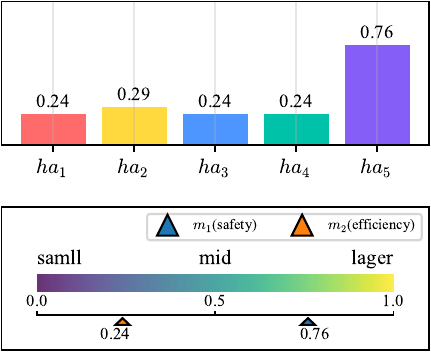}\label{fig: sub_figure5.9.1.b}}
  \caption{Status, reward weight, and mode for steps(70)}
  \label{fig5.9}
\end{figure}

To analyze the explainability of the proposed framework, one representative trajectory from task T4 is selected for visualization. The results are shown in Fig.~\ref{fig5.6}--Fig.~\ref{fig5.9}, corresponding to steps 0, 22, 61, and 73, respectively. In each figure, subfigure (a) illustrates the current environment state, where the red square denotes the target position, while subfigure (b) presents the corresponding reward weights ($ha_i$) and behavior mode values ($m_i$).

In the initial state, the robot is far from the target and remains stationary. From a human navigation perspective, a balanced strategy between safety and efficiency is generally preferred in such situations. Correspondingly, the learned mode is observed as $[0.46, 0.54]$, indicating a relatively balanced behavior tendency. Meanwhile, the reward weights $ha_1$--$ha_5$ remain at comparable levels, reflecting that the policy does not strongly bias either aggressive exploration or conservative avoidance.

In Fig.~\ref{fig5.7}, the robot has turned toward the target direction while approaching nearby obstacles. In human navigation understanding, turning behavior is expected in this situation, while obstacle proximity usually requires more conservative motion. Correspondingly, the turning-related reward weight $ha_2$ decreases from 0.68 to 0.35, reducing the penalty on steering behavior, whereas the collision-related weight increases to 0.73. The learned mode simultaneously shifts toward safety dominance, indicating consistency between the learned internal representation and human-understandable navigation behavior.

In Fig.~\ref{fig5.8}, the robot moves close to the target position, but its heading direction still deviates from the goal direction. In this situation, human navigation behavior generally emphasizes heading correction while maintaining movement efficiency in the absence of nearby obstacles. The observed results show that both the velocity-related reward weight $ha_1$ and the heading-related reward weight $ha_4$ increase to approximately 0.39, while the collision-related weight decreases. Correspondingly, the overall behavior mode shifts toward efficiency, indicating that the learned policy adapts its internal semantic representation according to environmental changes.

In Fig.~\ref{fig5.9}, the robot is about to reach the target but simultaneously moves closer to nearby obstacles. From a navigation safety perspective, more conservative behavior is typically required in this stage. Correspondingly, the safety-related mode component increases to 0.76, while the efficiency component decreases to 0.24, indicating that the policy dynamically adjusts its behavior preference to reduce collision risk near the goal region.

\begin{figure}[htbp]
\centering
\includegraphics[width=0.48\textwidth,height=0.12\textheight]{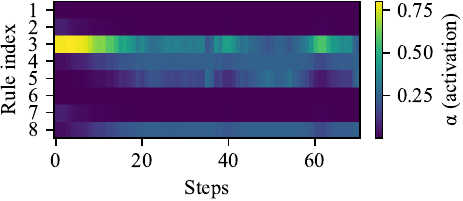}
  \caption{Rule activation heatmap}
  \label{fig5.10}
\end{figure}

Fig.~\ref{fig5.10} further visualizes the corresponding rule activation heatmaps. During the initial stage, Rule 3 dominates the inference process. In subsequent stages, the navigation conditions become more complex, and the policy behavior is jointly influenced by Rules 3, 4, 5, and 8. In particular, when the robot approaches obstacles, safety-related rules exhibit stronger activations, whereas efficiency-related rules become relatively weaker. Overall, the activated rules consistently exhibit a stronger tendency toward safety-oriented behavior. This visualization provides an explainable reasoning pathway from behavior modes to dynamic reward generation.

Overall, the results indicate that the learned behavior modes and reward weights maintain stable semantic correspondence with environmental changes and human navigation priors. Furthermore, the rule activation analysis demonstrates that the proposed framework provides an explainable reasoning mechanism rather than merely generating explainable final actions.

\subsection{Real-world Experiments}
To evaluate the performance of the proposed method in real-world scenarios, experiments were conducted in two real environments shown in Fig.~\ref{fig4.2.1}. Five target points were set in each environment (Table.\ref{tab5.4}), and the corresponding results are presented as follows.

\begin{table}[htbp]
\scriptsize
 \caption{{TARGET FOR EXPERIMENT 1 AND 2}}
 \label{tab5.4}
 \begin{center}
 \setlength{\tabcolsep}{3mm}{
 \begin{tabular}{ccccccc}
\toprule
 & Axes & \multicolumn{5}{c}{Target} \\
\midrule
\multirow{2}{*}{\textbf{Exp1}}
&$x$ & 1.20 & 1.40 & 1.60 &2.00& 2.50\\
&$y$ & 0.00 & 0.00 & 0.00 & 0.00& 0.00\\
\midrule
\multirow{2}{*}{\textbf{Exp2}}
&$x$ & 1.90 & 2.40 & 1.70 & 1.00& 0.20\\
&$y$ & -0.50 & 0.25 & 0.90 & 0.56& 0.00\\
\bottomrule
\end{tabular}}
\end{center}
\end{table}
\subsubsection{Experiment 1} 
\begin{figure}[ht]
\centering
\includegraphics[width=0.8\linewidth]{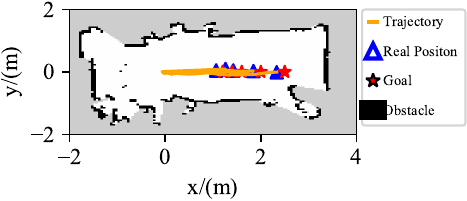}\\[-1pt]
\includegraphics[width=0.8\linewidth]{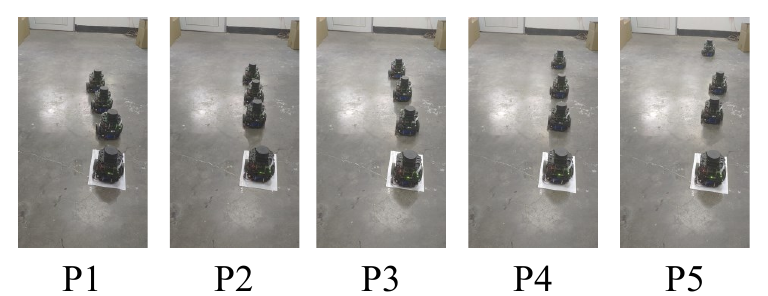}
\caption{Recorded trajectories and corresponding real image}
\label{fig5.11}
\end{figure}

This environment simulates an open-space scenario similar to Env1 and Env5. Fig.~\ref{fig5.11} presents the navigation trajectories and real-world motion snapshots. The robot exhibited behavior consistent with that observed during training, reaching all target points along relatively straight trajectories.

\begin{figure}[htbp]
\centering
\includegraphics[width=0.48\textwidth,height=0.11\textheight]{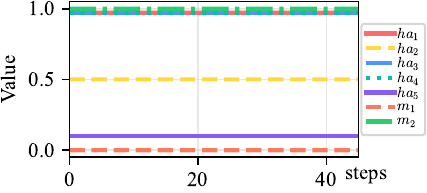}
  \caption{Mode and reward weight of Exp1 targert1}
  \label{fig5.12}
\end{figure}

Fig.~\ref{fig5.12} illustrates the variations of the reward weights and behavioral patterns during navigation toward the first target point. In the relatively open and safe environment, the behavioral pattern \(m_1\) remained at a low level, indicating that the policy tended to prioritize efficiency. Correspondingly, the collision-related reward weight \(ha_5\) was relatively small, while the velocity-related reward weight \(ha_1\) remained comparatively high. These observations suggest that the intermediate behavioral patterns can provide an explainable indication of the policy tendency during navigation.
\subsubsection{Experiment 2}
\begin{figure}[ht]
\centering
\includegraphics[width=0.8\linewidth]{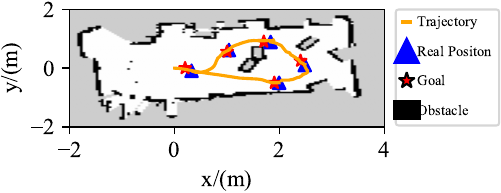}\\[-1pt]
\includegraphics[width=0.8\linewidth,height=0.2\linewidth]{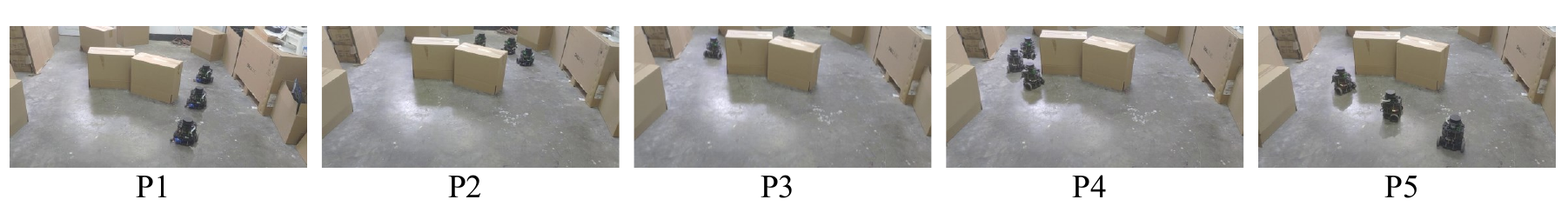}
\caption{Recorded trajectories and corresponding real image}
\label{fig5.13}
\end{figure}

This environment represents an unseen cluttered scenario with multiple obstacles, aiming to evaluate the generalization capability of the trained model. Fig.~\ref{fig5.13} presents the navigation trajectories and real-world motion snapshots. The robot successfully completed navigation tasks in the narrow multi-obstacle environment, indicating that the proposed method possesses a certain degree of generalization capability.

\begin{figure}[htbp]
\centering
\includegraphics[width=0.48\textwidth,height=0.12\textheight]{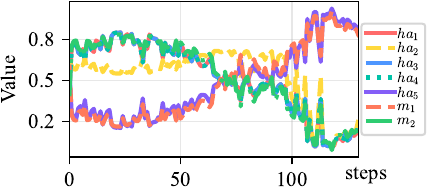}
  \caption{Mode and reward weight of Exp2 target1}
  \label{fig5.14}
\end{figure}

Fig.~\ref{fig5.14} shows the variations of the behavioral patterns and dynamic reward weights during navigation. Compared with the previous environment, the behavioral patterns in this scenario changed more dynamically. At the beginning of navigation, the environment was relatively open, and the policy adopted an efficiency-oriented mode. As the robot gradually approached surrounding obstacles, the safety-related pattern weights increased, causing the policy to shift toward a safety-oriented mode. Finally, when reaching the target position, the robot stopped and the behavioral patterns gradually became balanced.
Overall, the above results demonstrate the feasibility of deploying the proposed policy in real-world environments. In addition, the introduced intermediate behavioral patterns provide a certain degree of explainability for understanding the behavioral tendencies of the navigation policy.

\section{Conclusion}
This study investigated the integration of continual learning capability and policy explainability in robotic navigation tasks. Since continual learning prefers strong prior constraints while explainability-oriented behavior learning requires sufficient autonomous interaction, a conflict exists between the two objectives during policy optimization. To address this issue, an explainability-oriented continual reinforcement learning framework, FECHRL, was proposed. A two-stage behavior cloning mechanism was introduced to balance early-stage knowledge guidance and late-stage autonomous behavior formation, while a localized fuzzy critic was designed to stabilize the transition from cloning-guided learning to autonomous interaction learning. Experimental results in sequential navigation tasks demonstrated that the proposed framework can simultaneously maintain continual learning capability and generate semantically explainable behavior patterns. In particular, the proposed two-stage cloning strategy was shown to play an important role in balancing policy stability and explainability-oriented behavior learning.
Although the proposed method is effective, it still relies on multiple auxiliary mechanisms and hierarchical components, resulting in relatively high structural complexity. Future work will investigate whether similar sustained adaptation and explainability features can be achieved through more compact policy structures and intrinsic representation learning mechanisms.

\bibliographystyle{ieeetr}
\bibliography{my_references}

@article{1,
	title        = {A Reinforcement Learning Method With an Expert Guidance Mechanism for Manipulator Trajectory Generation},
	author       = {Pan, Qinghui and Lian, Jie and Liu, Xingjian and others},
	year         = 2026,
	journal      = {IEEE Transactions on Industrial Informatics},
	volume       = 22,
	number       = 3,
	pages        = {2068--2077},
	doi          = {10.1109/tii.2025.3634606}
}

@article{2,
	title        = {A goal-conditioned offline reinforcement learning algorithm and its application to quad-rotors},
	author       = {Haojun Zhong and Zhenlei Wang and Yuzhe Hao},
	year         = 2025,
	journal      = {Engineering Applications of Artificial Intelligence},
	volume       = 152,
	pages        = 110678,
	doi          = {https://doi.org/10.1016/j.engappai.2025.110678},
	issn         = {0952-1976}
}

@article{3,
	title        = {Behavior Cloning Assisted Reinforcement Learning for Cable-Driven Continuum Space Robots in Sparse Reward Environments},
	author       = {Tian, Xianru and Xia, Bo and Tan, Junbo and others},
	year         = 2025,
	journal      = {IEEE Robotics and Automation Letters},
	volume       = 10,
	number       = 9,
	pages        = {9232--9239},
	doi          = {10.1109/lra.2025.3588411}
}

@article{4,
	title        = {Close-range docking control for reconfigurable ground vehicles: Model-guided reinforcement learning with robust predictive safety filter},
	author       = {Xu Yang and Jun Ni and Hangjie Cen and others},
	year         = 2026,
	journal      = {Engineering Applications of Artificial Intelligence},
	volume       = 164,
	pages        = 113162,
	doi          = {https://doi.org/10.1016/j.engappai.2025.113162},
	issn         = {0952-1976}
}

@article{5,
	title        = {Curricular Subgoals for Inverse Reinforcement Learning},
	author       = {Liu, Shunyu and Qing, Yunpeng and Xu, Shuqi and others},
	year         = 2025,
	month        = mar,
	journal      = {Trans. Intell. Transport. Sys.},
	publisher    = {IEEE Press},
	volume       = 26,
	number       = 3,
	pages        = {3016-3027},
	doi          = {10.1109/tits.2025.3532519},
	issn         = {1524-9050},
	issue_date   = {March 2025},
	numpages     = 12
}

@article{6,
	title        = {Knowledge guided fuzzy deep reinforcement learning},
	author       = {Peng Qin and Tao Zhao},
	year         = 2025,
	journal      = {Expert Systems with Applications},
	volume       = 264,
	pages        = 125823,
	issn         = {0957-4174},

}

@article{7,
	title        = {Knowledge guided deep deterministic policy gradient},
	author       = {Qin, Peng and Zhao, Tao},
	year         = 2025,
	journal      = {Knowledge-Based Systems},
	publisher    = {Elsevier},
	pages        = 113087
}

@article{8,
	title        = {SurgIRL: Toward Life-Long Learning for Surgical Automation by Incremental Reinforcement Learning},
	author       = {Ho, Yun-Jie and Chiu, Zih-Yun and Zhi, Yuheng and others},
	year         = 2025,
	journal      = {IEEE Robotics and Automation Letters},
	volume       = 10,
	number       = 12,
	pages        = {13145-13152},
	doi          = {10.1109/lra.2025.3627088}
}

@article{9,
	title        = {SCRIPT: A Scalable Continual Reinforcement Learning Framework for Autonomous Penetration Testing},
	author       = {Shicheng Zhou and Jingju Liu and Yuliang Lu and others},
	year         = 2025,
	journal      = {Expert Systems with Applications},
	volume       = 285,
	pages        = 127827,
	doi          = {https://doi.org/10.1016/j.eswa.2025.127827},
	issn         = {0957-4174}
}

@article{10,
	title        = {Map-based experience replay: a memory-efficient solution to catastrophic forgetting in reinforcement learning},
	author       = {Hafez, Muhammad Burhan  and Immisch, Tilman  and Weber, Tom  and others},
	year         = 2023,
	journal      = {Frontiers in Neurorobotics},
	volume       = {17},
	doi          = {10.3389/fnbot.2023.1127642},
	issn         = {1662-5218}
}

@inproceedings{11,
	title        = {Experience Consistency Distillation Continual Reinforcement Learning for Robotic Manipulation Tasks},
	author       = {Zhao, Chao and Xu, Jie and Peng, Ru and others},
	year         = 2024,
	booktitle    = {2024 IEEE International Conference on Robotics and Automation (ICRA)},
	pages        = {501--507},
	doi          = {10.1109/icra57147.2024.10611494}
}

@article{12,
	title        = {AdaER: An adaptive experience replay approach for continual lifelong learning},
	author       = {Li, Xingyu and Tang, Bo and Li, Haifeng},
	year         = 2024,
	journal      = {Neurocomputing},
	publisher    = {Elsevier},
	volume       = 572,
	pages        = 127204
}

@article{13x,
  author={Qin, Peng and Zhao, Jianjian and Mei, Zhen and Yang, Hainan and Zhao, Tao},
  journal={IEEE Internet of Things Journal}, 
  title={Knowledge-Guided Continual Reinforcement Learning for Navigation}, 
  year={2026},
  volume={13},
  number={5},
  pages={9535-9547},
  }

@article{15,
	title        = {Optimal Formation Control for Autonomous Vehicles: A Bilayer Predefined-Time Fuzzy Reinforcement Learning Approach},
	author       = {Zhuang, Xinhai and Li, Hengyu and Liu, Jingyi and others},
	year         = 2025,
	journal      = {IEEE Transactions on Fuzzy Systems},
	volume       = 33,
	number       = 12,
	pages        = {4405--4417},
	doi          = {10.1109/tfuzz.2025.3613543}
}

@misc{16,
	title        = {Hierarchical Decision-Making for Autonomous Navigation: Integrating Deep Reinforcement Learning and Fuzzy Logic in Four-Wheel Independent Steering and Driving Systems},
	author       = {Yizhi Wang and Degang Xu and Yongfang Xie and others},
	year         = 2025,
	eprint       = {2508.16574},
	archiveprefix = {arXiv},
	primaryclass = {cs.RO}
}

@article{17,
	title        = {Knowledge guided self-evolving fuzzy deep reinforcement learning with dual-adaptive optimization strategy},
	author       = {Suyu Zhu and Mingjun Li and Dean Hu and others},
	year         = 2026,
	journal      = {Expert Systems with Applications},
	volume       = 307,
	pages        = 131004,
	doi          = {https://doi.org/10.1016/j.eswa.2025.131004},
	issn         = {0957-4174}
}

@article{18,
	title        = {Reinforcement Learning-Based IT-2 Fuzzy Fractional-Order Sliding Mode Control for Leader-Follower Formation Tracking of Flexible-Joint Robots},
	author       = {Narayanan, G. and Lee, Sangmoon and Ahn, Sangtae},
	year         = 2026,
	journal      = {IEEE Transactions on Fuzzy Systems},
	volume       = 34,
	number       = 2,
	pages        = {503--516},
	doi          = {10.1109/tfuzz.2025.3639438}
}

@article{19,
	title        = {Fuzzy Reinforcement Learning for Adaptive Control of CDQ Systems in Steel Industry},
	author       = {Zhang, Han and Liu, Yang and Zhao, Jun and others},
	year         = 2026,
	journal      = {IEEE Transactions on Industrial Informatics},
	pages        = {1--12},
	doi          = {10.1109/tii.2026.3671728}
}

@article{20,
	title        = {Predictor-Based Event-Triggered Optimized Control for Uncertain Nonlinear Systems With Time Delays: A Reinforcement Learning Approach},
	author       = {Li, Ping and Fu, Li and Song, Zhibao and others},
	year         = 2025,
	journal      = {IEEE Transactions on Fuzzy Systems},
	volume       = 33,
	number       = 9,
	pages        = {3360--3374},
	doi          = {10.1109/tfuzz.2025.3592833}
}

@article{21,
	title        = {Evaluating reinforcement learning-based neural controllers for quadcopter navigation in windy conditions},
	author       = {Alain Andres and Aritz D. Martinez and Sümer Tunçay and others},
	year         = 2025,
	journal      = {Engineering Applications of Artificial Intelligence},
	volume       = 161,
	pages        = 112090,
	doi          = {https://doi.org/10.1016/j.engappai.2025.112090},
	issn         = {0952-1976}
}

@article{22,
	title        = {DAG-NAS : An explainable neural architecture search framework for reinforcement learning},
	author       = {Taegun An and Changhee Joo},
	year         = 2026,
	journal      = {Neural Networks},
	volume       = 201,
	pages        = 108901,
	doi          = {https://doi.org/10.1016/j.neunet.2026.108901},
	issn         = {0893-6080}
}

@article{23,
	title        = {Interactive Autonomous Navigation With Internal State Inference and Interactivity Estimation},
	author       = {Li, Jiachen and Isele, David and Lee, Kanghoon and others},
	year         = 2024,
	journal      = {IEEE Transactions on Robotics},
	volume       = 40,
	pages        = {2932--2949},
	doi          = {10.1109/tro.2024.3400937}
}

@article{24,
	title        = {A curriculum-guided and explainable reinforcement learning framework for fixed-wing unmanned aerial vehicle autopilots},
	author       = {Yunxiao Lian and Ni Li and Pan Zhou and others},
	year         = 2026,
	journal      = {Engineering Applications of Artificial Intelligence},
	volume       = 163,
	pages        = 113015,
	doi          = {https://doi.org/10.1016/j.engappai.2025.113015},
	issn         = {0952-1976}
}

@article{25,
	title        = {FinQuaXBot: enhancing trust and security in personalized investment and tax forecasting using homomorphic encryption and meta-reinforcement learning with explainability},
	author       = {I. Jency and R.P. {Anto Kumar}},
	year         = 2025,
	journal      = {Expert Systems with Applications},
	volume       = 287,
	pages        = 128136,
	doi          = {https://doi.org/10.1016/j.eswa.2025.128136},
	issn         = {0957-4174}
}

@article{ewc,
	title        = {Overcoming catastrophic forgetting in neural networks},
	author       = {James Kirkpatrick  and Razvan Pascanu  and Neil Rabinowitz  and others},
	year         = 2017,
	journal      = {Proceedings of the National Academy of Sciences},
	volume       = 114,
	number       = 13,
	pages        = {3521--3526},
	url          = {https://www.pnas.org/doi/abs/10.1073/pnas.1611835114},
	eprint       = {https://www.pnas.org/doi/pdf/10.1073/pnas.1611835114}
}

@inproceedings{si,
	title        = {Continual learning through synaptic intelligence},
	author       = {Zenke, Friedemann and Poole, Ben and Ganguli, Surya},
	year         = 2017,
	booktitle    = {Proceedings of the 34th International Conference on Machine Learning},
	location     = {Sydney, NSW, Australia},
	
	volume       = 70,
	pages        = {3987--3995},
	numpages     = 9
}

@inproceedings{CORESETS,
	title        = {Selective experience replay compression using coresets for lifelong deep reinforcement learning in medical imaging},
	author       = {Zheng, Guangyao and Zhou, Samson and Braverman, Vladimir and others},
	year         = 2024,

	booktitle    = {Medical Imaging with Deep Learning},
	
	volume       = 227,
	pages        = {1751--1764},
	url          = {https://proceedings.mlr.press/v227/zheng24a.html}
}

@inproceedings{SCDMSLL,
	title        = {Sparse coding in a dual memory system for lifelong learning},
	author       = {Sarfraz, Fahad and Arani, Elahe and Zonooz, Bahram},
	year         = 2023,
	booktitle={Proceedings of the AAAI Conference on Artificial Intelligence},
    volume={37},
    number={8},
	pages={9714--9722},
	isbn         = {978-1-57735-880-0},
	url          = {https://doi.org/10.1609/aaai.v37i8.26161},
	articleno    = 1092,
}

@article{ddpg,
      title={Continuous control with deep reinforcement learning}, 
      author={Timothy P. Lillicrap and Jonathan J. Hunt and Alexander Pritzel and Nicolas Heess and Tom Erez and Yuval Tassa and David Silver and Daan Wierstra},
      year={2019},
      eprint={1509.02971},
      archivePrefix={arXiv},
      primaryClass={cs.LG},
      url={https://arxiv.org/abs/1509.02971}, 
}

@ARTICLE{5X,
  author={Qin, Peng and Zhao, Jianjian and Zhao, Jiayu and Zhao, Tao},
  journal={IEEE Transactions on Neural Networks and Learning Systems}, 
  title={Guided Continual Reinforcement Learning for Navigation via Integrative Consolidation and Memory Augmentation}, 
  year={2026},
  volume={},
  number={},
  pages={1-15},
  doi={10.1109/TNNLS.2026.3746646}}

\end{document}